## ERRATUM

# Link prediction in drug-target interactions network using similarity indices

Yiding Lu, Yufan Guo and Anna Korhonen[*]

[*]Correspondence:
alk23@cam.ac.uk
Computer Laboratory, University
of Cambridge, JJ Thompson
Avenue, Cambridge, UK
Full list of author information is
available at the end of the article

Dr. Carlo Vittorio Cannistraci has kindly suggested that the modified common neighbors (CN) index reported in this paper – when applied to unweighted networks – is interchangeable with the local community links (LCL) index proposed in (Daminelli et al., 2015). After a detailed discussion on the definition of "quadrangular closure" introduced in (Daminelli et al., 2015), particularly a clarification from the authors about the statement "the quadrangular closure encloses the shortest path between two non-adjacent nodes that belong to two distinct classes", we can confirm that this is indeed the case. The comparative analysis against state-of-the-art link prediction methods with bipartite network projection as reported in (Daminelli et al., 2015), along with the comparison between our approaches - purely based on network topology - against those leveraging additional information about the characteristics of drugs, targets and DTIs as reported in this paper and in Daminelli et al, demonstrate that variations of CN index adapted for bipartite networks have a great potential in DTI prediction.

**Reference**: Simone Daminelli et al. 2015. Common neighbours and the local-community-paradigm for link prediction in bipartite networks. New J. Phys. 17 113037

 

# Link prediction in drug-target interactions network using similarity indices

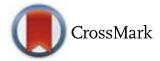

Yiding Lu, Yufan Guo and Anna Korhonen*

## Abstract

**Background:** *In silico* drug-target interaction (DTI) prediction plays an integral role in drug repositioning: the discovery of new uses for existing drugs. One popular method of drug repositioning is network-based DTI prediction, which uses complex network theory to predict DTIs from a drug-target network. Currently, most network-based DTI prediction is based on machine learning – methods such as Restricted Boltzmann Machines (RBM) or Support Vector Machines (SVM). These methods require additional information about the characteristics of drugs, targets and DTIs, such as chemical structure, genome sequence, binding types, causes of interactions, etc., and do not perform satisfactorily when such information is unavailable. We propose a new, alternative method for DTI prediction that makes use of only network topology information attempting to solve this problem.

**Results:** We compare our method for DTI prediction against the well-known RBM approach. We show that when applied to the MATADOR database, our approach based on node neighborhoods yield higher precision for high-ranking predictions than RBM when no information regarding DTI types is available.

**Conclusion:** This demonstrates that approaches purely based on network topology provide a more suitable approach to DTI prediction in the many real-life situations where little or no prior knowledge is available about the characteristics of drugs, targets, or their interactions.

## Background

*In silico* prediction of drug target interactions (DTIs) refers to an automated search for potential interactions between chemicals and proteins. DTI prediction plays an integral role in drug repositioning, i.e., in the discovery of new uses for existing drugs. Candidates for drug repositioning have typically undergone several stages of clinical development and therefore have well-known safety and pharmacokinetic profiles. This allows accelerating pharmaceutical research and development without increasing the risk of failed developments [1]. Drug repositioning can be facilitated through DTI prediction that discovers likely interactions between chemicals and proteins, allowing researchers to identify potential new uses of drugs. For example, the drug duloxetine was originally developed for depression. It works by blocking the reuptake of serotonin and noradrenaline in the synaptic cleft. However, serotonin and noradrenaline also exert an excitatory effect on urethral sphincter motor neurons, protecting against urine leakage. Therefore, duloxetine was repositioned to stress urinary incontinence treatment which was proven effective in clinical trials. Another successful case of drug repositioning is dapoxetine, which was developed to cure analgesia but was later used for premature ejaculation.

Many social, biological and information systems can be represented in the form of networks, where nodes represent concepts such as individuals, chemicals, proteins and web users, and edges represent the interactions or relationships between the nodes [2]. In a DTI network, we have a bipartite graph, with drugs and targets forming two disjoint sets of nodes and the interactions between the drugs and targets forming the edges. Such complex networks are assumed to exhibit organising principles that should be contained partially in the network topology. This has led to the birth of network analysis, whereby researchers try to develop tools and measurements to discover the latent organising principles of complex networks.

One important aspect of network analysis is link prediction. This involves estimating the likelihood of the existence of a link between two nodes, based on the observed links in the network and the characteristics of nodes and links [3]. In our paper, we focus on the application of link prediction to DTIs - an emerging approach starting to gain more attention in the past decade [4–7]. Just one recent example is the work of [4] that uncovered the potential interaction between spironolactone and membrane progestin receptor gamma protein, and that has been confirmed in recent clinical studies [8].

*Correspondence: alk23@cam.ac.uk
Computer Laboratory, University of Cambridge, JJ Thompson Avenue, Cambridge, UK

Most current network-based methods for DTI prediction are based on machine learning, such as the work of [4] based on restricted Boltzmann machines (RBM), and that of [7] based on support vector machines (SVM), among others [9, 10]. Although such methods tend to perform well, they require prior information about the characteristics of proteins and chemicals. For example, the model in [11] uses chemical structure and protein sequence similarity as basis of its classification rules, and the model in [4] produces less satisfactory results when the modes of interaction between chemicals and proteins are unknown. This is problematic because databases containing DTIs might only be partly annotated or even unannotated. For example DrugBank, one of the largest resources for drugs and proteins, only provides annotations on the targets but not the interactions. The Therapeutic Target Database, on the other hand, does not link targets to protein databases making it difficult to retrieve information such as protein structure [12]. Moreover, as concluded in [9], DTI prediction based on similarity between chemical structures or protein sequences has limitations since its underlying assumption that similar drugs share similar targets is not necessarily true. Also, the large portion of so-called "Me Too" drugs might have made the evaluation of such methods too optimistic.

We therefore propose a new method for DTI prediction purely based on DTI network topology, without accessing external knowledge other than known DTIs, and with a special focus on high precision for top-ranked predictions. It employs similarity indices that have been used for link prediction in various complex networks [2, 13].

Two nodes are considered similar if they have common features [14].

We focus on the topological similarity between nodes which is based solely on the network structure [2].

The algorithm we employ is simple to implement. A score $s_{xy}$ is assigned to each pair of nodes $x$ and $y$ based on the similarity between $x$ and $y$. All resultant non-observed links are then ranked according to their score and the links connecting more similar nodes have a higher likelihood of existence. While there have been very few recent works on DTI prediction purely based on network topology information [5, 6], our work differs from those approaches in that it ranks all candidate drug-target pairs together, instead of ranking predicted drugs for each individual target [5] or providing an unsorted list of novel drug use suggestions [6].

Although similarity indices have been used for link prediction in social networks with promising results [15], they have not, to the best of our knowledge, been used to predict links in DTI networks. In this paper, we investigate whether the method is applicable to this task. We apply four types of similarity indices to the Manually Annotated Target and Drugs Online Resources (MATADOR) database of DTIs [12]: Common Neighbours, Jaccard index, Preferential Attachment and Katz index, after first adapting them (where needed) to the bipartite nature of the DTI network.

While the selected indices look simple, they are demonstrated to be the most robust similarity-based algorithms with superior performance on complex networks [2, 15], and significantly outperform sophisticated machine learning-based approaches (RBM, [4]) on DTI networks when no finer-grained annotation about DTI types is available. We demonstrate that when the modes of interaction between drugs and targets are unspecified, Common Neighbours, Jaccard index and Katz index have significantly higher precision than RBM for the top 5000 predicted links. This is important as this provides researchers with more reliable potential DTIs for further experimental validation, saving time and money. Our new method is preferable in situations where the input dataset does not have a comprehensive coverage of entity/interaction subtypes or entity similarities for all proteins or chemicals. When the input data is partially annotated, it could be used as a powerful complementary method alongside the conventional machine learning -based methods that tend to perform well on annotated datasets.

## Method

### MATADOR database
The Manually Annotated Target and Drug Online Resource (MATADOR) database is a free online database of DTIs [12]. It includes all modes of interaction between chemicals and proteins, unlike other resources such as DrugBank which only includes the main modes of interaction. These modes of interaction between drugs and targets can be direct, indirect, and in some cases, combination of the two.

The MATADOR database, as of April 2015, contains a total of 15,843 DTI entries. Each entry consists of 13 fields. In our work we only use 3 of the fields - those that are most relevant for our experiments:

1. Chemical ID: The PubChem compound identifier.
2. Protein ID: The protein identifier, either corresponding to genes from STRING 7 database or from Medical Subject Headings (MeSH)

3. MATADOR Score: A score for the confidence in the interaction.

The Chemical and Protein IDs in each entry of the database are used for forming the DTI network. For example, if the Chemical ID is 1 and the Protein ID 50, we connect node 1 in the set of chemicals to node 50 in the set of proteins. The MATADOR score allows us to create a weighted version of the DTI network.

### Graph representation of DTI network

A DTI network can be represented as a graph, where drugs and targets are the nodes and DTIs are the edges. This can be represented by $G = (V, E)$ where $V$ is the set of nodes and $E$ is the set of edges. We can picture a graph by using dots as nodes and lines as edges. A common method to represent a graph is an adjacency matrix. An adjacency matrix is a $n \times n$ matrix where $n$ represents the number of nodes in the network and the entries $a_{ij}$ are given by [16]:

$$a_{ij} = \begin{cases} 1 & \text{if there is an edge between } i \text{ and } j \\ 0 & \text{otherwise} \end{cases}$$

The network edges can also be weighted. In this case, instead of $a_{ij} = 1$ when there is an edge between nodes $i$ and $j$, we use the weight of the edge instead.

The DTI network formed from the MATADOR database is a bipartite graph. This means that the set of nodes $V$ can be partitioned into two disjoint sets $U$ and $W$ whereby each edge in the network links one node from $U$ to another one in $W$ and there are no edges between nodes within $U$ or $W$. In our case, the two sets of nodes are chemicals and proteins. Since there are no interactions within the MATADOR database which depict chemical-chemical interactions or protein-protein interactions, we form a bipartite graph and represent the network in the form of a biadjacency matrix. The creation process is similar to the adjacency matrix except that the matrix is $n \times m$, where $n$ represents the number of nodes in set $U$ and $m$ the number of nodes in set $W$.

### Similarity indices

Similarity-based algorithms can be considered as the simplest framework for link prediction in networks. As mentioned earlier, they are based on the theory that two nodes are more likely to have a link between them if they share many common features. One of these common features is the common neighbours which they share, which can be calculated solely based on the network structure. This ensures that similarity indices can easily be implemented on networks even when no prior knowledge of the nodes is available. Similarity indices can be derived from local or global network topology. Local indices are node-dependent, which means that the only information required are the degrees of the node and its nearest neighbourhood. Global similarity indices, in contrast, are path-dependent and global knowledge of the network topology is required [17].

Similarity indices have proved promising for link prediction in many different networks. Liben-Nowell and Kleinberg (2007) used a variety of them to predict the evolution of social networks over time. They showed that many similarity indices, such as Common Neighbours and Jaccard index, vastly outperform a random link predictor, indicating that a lot of information about networks is contained within the network topology alone. Zhou, Lü and Zhang (2009) applied several similarity indices to 6 different networks and obtained a high level of area under precision-recall curve (AUPR) for the majority of them. Given these promising results we believe that similarity indices could similarly be applied to DTI networks to identify potential chemical-protein interactions.

One challenge with using similarity indices in DTI networks is that most indices were originally proposed for social networks which are not bipartite in nature, and cannot be directly applied to a bipartite DTI network. For example, in Common Neighbours, if we try to predict a link between chemical $c_i$ and protein $p_j$, we will find that the neighbouring nodes of $c_i$ are all proteins and likewise, the neighbouring nodes of $p_j$ are all chemicals. Therefore, the simple Common Neighbours index will give us zero value for all $c_i$ and $p_j$.

However, there are other type of networks which are bipartite in nature. One example is a user recommendation network where the two disjunct sets are users and recommended items [18]. Modified similarity indices have been developed for such bipartite graphs. We will explain these further in the following sections where we introduce the four types of similarity indices used in our work, selected on the basis of their successful use in related work on other types of network [13, 17, 19].

### Common neighbours (CN)

For a node $x$, let $\Gamma(x)$ denote the set of neighbours of $x$. With this in mind, if two nodes $x$ and $y$ share many common neighbours, a link is likely to exist between these two nodes. A simple measure of this is given by:

$$S_{xy}^{CN} = |\Gamma(x) \cap \Gamma(y)|$$

This essentially counts the number of nodes which have both $x$ and $y$ as their neighbouring nodes. As mentioned previously, in a bipartite network, if we look at chemical $x$, its neighbours will always be proteins. At the same time, there are no links between proteins and proteins. This means that $|\Gamma(x) \cap \Gamma(y)|$ will always be zero. Hence, we need to modify our definition of CN for a bipartite graph. Now, we define $\widehat{\Gamma}(x) = \bigcup_{c \in \Gamma(x)} \Gamma(c)$ as the set of neighbours of node $x$'s neighbours [20], we can then redefine CN as:

$$S_{xy}^{CN'} = |\Gamma(x) \cap \widehat{\Gamma}(y)|$$

In the original CN index, we basically count the total number of unique paths of length 2 ($x$ to a common neighbour, common neighbour to $y$). In the modified CN index, we have increased the path length to 3. Thus, we can represent CN in the following way:

$$S_{xy} = \sum_{\substack{z_1 \in \Gamma(x) \cap \Gamma(z_2) \\ z_2 \in \Gamma(z_1) \cap \Gamma(y)}} w(x, z_1) + w(z_1, z_2) + w(z_2, y)$$

For an unweighted network, all 3 values of $w$ are 1. For a weighted network, we use the weight of each link in the path instead.

### Jaccard index
The Jaccard index is a commonly used similarity metric in information retrieval. For a randomly selected feature $f$ of either node $x$ or node $y$, the Jaccard index measures the probability that both nodes $x$ and $y$ have feature $f$ [15]. In our case, the features are the neighbours of the node; therefore we define Jaccard index as:

$$S_{xy}^{Jaccard} = \frac{|\Gamma(x) \cap \Gamma(y)|}{|\Gamma(x) \cup \Gamma(y)|}$$

Similar to CN, we have to modify the Jaccard index for a bipartite graph [20]:

$$S_{xy}^{Jaccard'} = \frac{|\Gamma(x) \cap \widehat{\Gamma}(y)|}{|\Gamma(x) \cup \widehat{\Gamma}(y)|}$$

The Jaccard index [21] is basically a normalised version of CN, taking into account the influence of a node in the network. For instance, in a social network, a highly influential individual is naturally well-connected to other individuals in the network. Therefore, it is likely that two highly influential individuals will share many common neighbours even though they are not close friends and the majority of their friends do not overlap at all. In this case, they will obtain a high CN score based on their influence in the network. The Jaccard index solves this problem by placing more emphasis on the links of non-influential nodes to ensure that the common neighbours they share are due to their similarity rather than their influence.

For the weighted Jaccard index, we simply take the weighted CN and divide it by the total number of neighbours between nodes $x$ and the neighbouring nodes of $y$.

### Preferential attachment (PA)
Let $k_x$ be the degree of node $x$, then PA between node $x$ and $y$ is defined as [2]:

$$S_{xy}^{PA} = k_x \times k_y$$

PA is based on the phenomenon that nodes with many links tend to generate more new links. This phenomenon can be found in many scenarios. For example, film actors who are well-connected in Hollywood are more likely to acquire new roles in movies which then increase their fame [22]. Likewise, in scientific journals, the most cited articles induce researchers to read them and hence increase their citation numbers. This is known as the Matthew effect [23] where the "rich gets richer".

PA does not require information about the neighbourhood and is only dependent on the degree of the nodes $x$ and $y$. It has the lowest computational complexity of all the similarity indices, and does not require any modification of the bipartite graph. For weighted PA, instead of using the degree of the node $x$, we use the sum of the weights between node $x$ and its neighbours, therefore we have the following index:

$$S_{xy}^{PA} = \sum_{z_1 \in \Gamma(x)} w(x, z_1) \times \sum_{z_2 \in \Gamma(y)} w(z_2, y)$$

### Katz index
The Katz index [24] is a path-dependent global similarity index, which directly sums over the collection of paths between two nodes in a network and is exponentially damped to give the shorter paths more weight [2]. Let $A$ be the adjacency matrix whereby $a_{xy} = 1$ if $x$ is connected to $y$, else $a_{xy} = 0$; The Katz index can be defined as:

$$s_{xy}^{Katz} = \beta A_{xy} + \beta^2 (A^2)_{xy} + \beta^3 (A^3)_{xy} + \dots$$

The whole similarity matrix can be written as:

$$S^{Katz} = (I - \beta A)^{-1} - I$$

The damping factor $\beta$ controls the path weights. A small $\beta$ value means that longer paths contribute less to the Katz index score and vice versa. This means that small $\beta$ values yield results similar to CN. To ensure that the Katz index converge, the value of $\beta$ must be less than the reciprocal of the largest eigenvalue of $A$. Since the Katz index requires the calculation of an inverse of the matrix, we use the adjacency matrix $A$ created from the biadjacency matrix $B$ using the method explained in Section "Baseline". This works for both weighted and unweighted biadjacency matrix $B$.

### Implementation
The programming language and development environment used in our experiment and analysis of results is MATLAB R2014b. This was chosen as we represent DTI networks in the form of matrices and MATLAB has its own built-in libraries for manipulating and calculating matrices. Moreover, MATLAB supports the development of applications with graphical user interfaces (GUI) via graph-plotting tools. The function `plot` can plot a graph from two input vectors $x$ and $y$. This facilitates plotting the precision-recall (PR) curve of the results obtained after applying the similarity indices and to and compare the different indices in terms of performance.

The entries of the MATADOR database were stored in an Excel file. This was first read into MATLAB using the command `xlsread`. Following that, we could form the biadjacency matrix using the various fields of the MATADOR entries as explained in Section "MATADOR database". In doing so, we obtained a $2901 \times 801$ matrix $\boldsymbol{B}$ with 15843 non-zero entries.

### Evaluation
#### 10-Fold cross validation
Cross validation is a common technique for assessing whether the results of the similarity indices will generalise to any independent dataset. One of the most popular cross validation methods is the $k$-fold cross validation: the input dataset was partitioned into $k$ sub-datasets of approximately equal size. The experiment was then performed $k$ times where each time, one out of the $k$ sub-datasets was used as validation data and the other $k - 1$ sub-datasets were used as training data. The $k$ results were then averaged to obtain a single result.

For our experiments, we used 10-fold cross validation, which is known to give the lowest bias and variance in the sub-datasets [25]. Hence, the entries within the MATADOR database were randomly divided into 10 non-overlapping subsets of approximately equal size in terms of the number of DTIs. Following that, we created a biadjacency matrix for each of the 10 subsets. For each similarity index, we applied the algorithm to the sum of 9 biadjacency matrices and used the remaining biadjacency matrix as the validation data to check if the links that we predicted were correct. This process was repeated 10 times, and the precision and recall calculated from each iteration were averaged to produce a final score.

#### Precision-recall curve
Precision is defined as:

$$\text{Precision} = \frac{\text{True Positive}}{\text{True Positive} + \text{False Positive}}$$

and recall as:

$$\text{Recall} = \frac{\text{True Positive}}{\text{True Positive} + \text{False Negative}}$$

In our case of link prediction, true positive (TP) refers to the links predicted using the training data that are found in the validation data. False positive (FP) refers to the links predicted using the training data that are not found in the validation data. False negative (FN) refers to the links that are not predicted using the training data but are found in the validation data.

In our experiments, after applying a similarity index to the training data, we ranked the links predicted according to their scores. Then, we took the top $n$ links predicted and calculated the precision and recall based on the $n$ links. This was repeated for $n = 1$ to 10000. With this, we obtained 10000 points at different precision and recall values. This is then averaged for the 10 iterations of the different training data for each similarity index. We could then plot the PR curve for each similarity index and compare their performance.

### Baseline - restricted boltzmann machine (RBM)
We compared our method against the recent work of [4], who used RBM for DTI prediction and demonstrated good performance on the MATADOR database.

An RBM is a two-layer graphical model consisting of one layer of "visible" units, or observed states; and one layer of "hidden" units, or feature detectors [26]. In DTI prediction, an RBM can be created for each target, where each visible unit represents the characteristics of interaction between the target and a certain drug. In [4]'s model, each visible unit is composed of two variables $x_{direct}$ and $x_{indirect}$, indicating whether a target and a drug has direct/indirect interactions. The RBMs for all the targets share the same parameters between hidden and visible layers. They can be trained and used for DTI prediction in the same fashion as in collaborative filtering [27].

## Results
### Baseline
According to [4], the RBM model was tested in the following three scenarios. The results are as illustrated in Fig. 1.

1. Integrating both direct and indirect DTIs with distinction, the input "visible" unit is a multidimensional vector indicating the mode of interaction
2. Mixing both direct and indirect DTIs without distinction; input "visible" unit is a one-dimensional binary vector indicating whether DTIs are observed
3. Using only a single interaction type

Despite the remarkable performance of RBM when trained on data that distinguish between direct and indirect links, when mixing the two types of links, the results become less satisfactory. Especially in the prediction of indirect DTIs using RBM, we can see that the precision is only around 0.4. This is mainly due to the RBM using the mode of interaction as a feature when predicting DTIs. When mode of interaction is available, the input unit is multidimensional. However, when the mode of interaction is not available, in the input unit is one-dimensional, which greatly lowers the precision of the RBM method.

Wang and Zeng kindly provided the results data for their paper using the RBM method [4]. Using the results data,

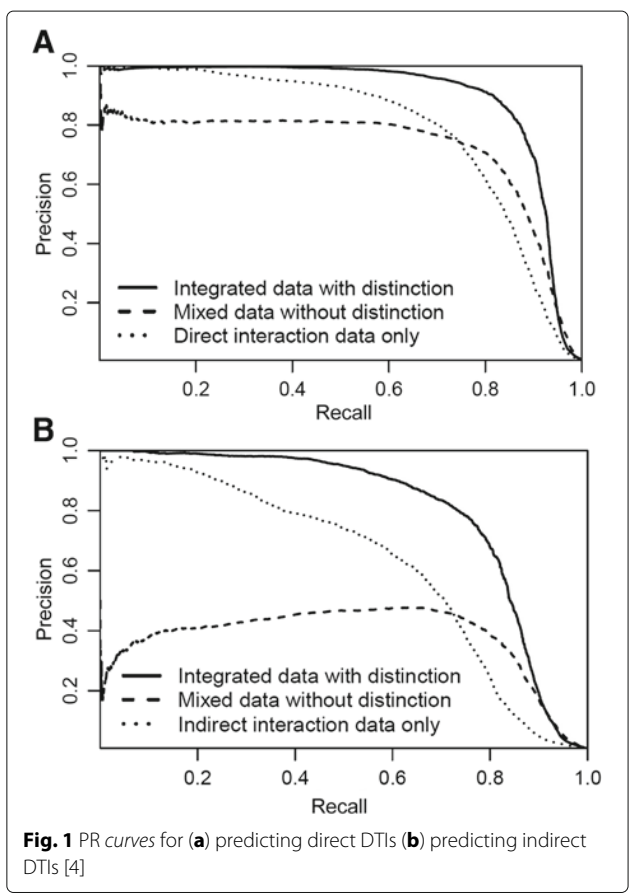

**Fig. 1** PR *curves* for (**a**) predicting direct DTIs (**b**) predicting indirect DTIs [4]

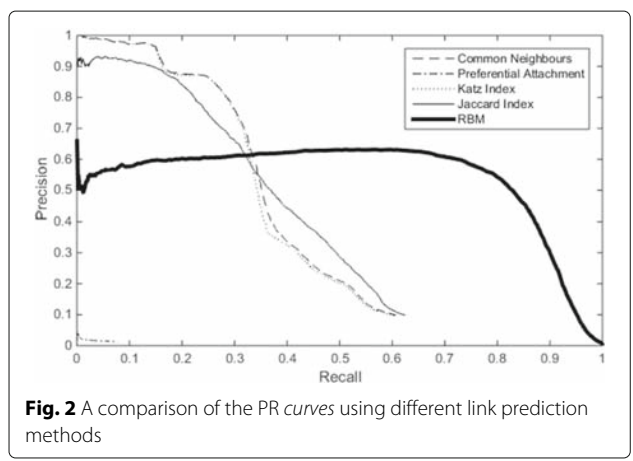

**Fig. 2** A comparison of the PR *curves* using different link prediction methods

we are able to plot a PR curve which combines the result of predicting direct and indirect DTIs This is the "baseline" method we compare the different similarity indices against.

### Similarity indices

To obtain a fair comparison with their experiment, we focus on cases where no distinction between the links in the DTI network is made and therefore, we use only the unweighted version of the similarity indices. We chose the $\beta$ value for Katz Index to be 0.005 as it has the best performance out of the 3 different values (0.005, 0.01, 0.02).

The results are as shown in Fig. 2. We can see that three of the four similarity indices (CN, Jaccard index and Katz index) have significantly higher precision than baseline when recall is between 0 and 0.35, and interestingly their PR curves intersect almost at the same point. Hence, there is a common cost ratio threshold for determining the better-performing methods whose top-rank predictions are more accurate. While in terms of AUPR, similarity index-based approaches do not outperform RBM (Table 1), the fact that their PR curves intersect does show that there is no certain method dominating the other, and that one method is better for some cost ratios and worse for others. The "area under curve" measure, as reported in [28], has limitations and could be a misleading measure of performance in certain scenarios. Particularly, it weights omission and commission errors equally, and summarises test performance over regions in which one would rarely operate. In the scenario of drug repurposing, where any predicted links need to be validated through expensive, time-consuming laboratory experiments and clinical trials, false positives and false negatives do not have the same effect. It is critical to ensure that top-ranked predictions are highly reliable in the interest of time, money, and risk control. Our method is therefore better suited than the RBM approach in situations where no information is available about the modes of interaction between chemicals and proteins.

### Discussion

Taking a closer look at the trends of PR curve for the various similarity indices, first, we notice that Katz index and CN have very similar PR curves. This is because we are using a very small value of $\beta$, which gives more weight to shorter paths and less weight to longer ones. Thus, the Katz index functions almost like CN, producing a similar PR curve. We assume that the Katz index should have a larger maximum recall than CN, as it is able to predict links in the validation dataset that require more than 3 steps of the links in the training dataset to reach. Indeed, when we look at the total number of links predicted by each link prediction method, CN predicts around 580,000 links while Katz index predicts about 1.85 M. As this is more than 3 times the number predicted by CN, Katz index is expected to have better recall. However, looking at the PR curve for both methods, the maximum recall the two indices is about 0.6. This shows that the top 10000 links predicted are not actually affected by the longer paths incorporated in Katz index. If we look at the PR curves closely, CN has the best performance out of the 3 similarity indices at low recall. This is unexpected as CN is the simplest index, and intuition tells us that the more complex indices should have better performance as they have factored in more information about network structure. However, this is not the case with the DTI network. In fact, this has also been observed by various other researchers. Zhou, Lü and Zhang (2009) used 9 different similarity indices for link prediction on 6 different networks and found that CN performed the best for all the datasets [13]. This shows that the CN actually is a very strong link prediction method despite its simplicity.

**Table 1** AUPR for baseline and similarity index-based approaches

|  | AUPR |
| --- | --- |
| Baseline | 0.5398 |
| Common Neighbours | 0.3715 |
| Jaccard Index | 0.3697 |
| Preferential Attachment | 0.0022 |
| Katz Index ($\mu = 0.005$) | 0.3652 |
| Katz Index ($\mu = 0.01$) | 0.3486 |
| Katz Index ($\mu = 0.02$) | 0.3030 |

Comparing CN, Katz index against Jaccard index, the former has a sharp drop in precision at about 0.15 recall and also between 0.3 to 0.35 recall. This is mainly because a large number of predicted links end up with equal similarity score according to the same number of common neighbours. Looking at Jaccard index, however, its precision decreases at a slower rate and it has a much smoother curve from the effect of its normalising factor. Although for recall below 0.35, Jaccard index does not perform as well as CN or Katz index, as recall increases, it does demonstrate better performance than other indices. Yet, as explained above, it is still more important to have higher precision especially when recall is low. Therefore, Katz index and CN are more suitable for our task than the Jaccard index.

PA shows the lowest performance out of all methods. Its poor performance is probably because it is originally designed based on the influence level of individuals in social networks. As previously mentioned in Preferential attachment section, PA is based on the Matthew effect (e.g., a new member of a social network is much more likely to know about an influential member of the network than about a non-influential member and, therefore, the influence of a person in the network further increases his influence level). However, in a DTI network, it turns out that even if a protein interacts with many different chemicals, it does not necessarily make the protein more likely to interact with the next chemical. It seems that the underlying assumption of PA does not work in a DTI network causing it to have poor performance.

## β values in Katz index

We further investigated the impact of different $\beta$ values on the PR curve of Katz Index for DTI prediction. Figure 3 below shows the 3 different PR curves using $\beta$ values of 0.005, 0.01 and 0.02 for Katz Index. It is clear that a smaller $\beta$ leads to better performance than a larger $\beta$. To explain this effect, we conducted an additional experiment on one of the training datasets. Assuming two nodes $x$ and $y$ share a link in the validation dataset, we would like to find out the length of path between $x$ and $y$ using the links in the training dataset. For example, as shown in Fig. 3, to reach from node $x$ to node $y$ (a link present in the validation dataset), we need to move from $x$ to $a$, $a$ to $b$ and finally $b$ to $y$. This shows that a path length of 3 is required using the links in the training dataset to reach the link in the validation dataset.

The aim of this experiment was to determine the average path length required in the training data to reach the links in the validation data. Since the input dataset is partitioned into training dataset and validation dataset, we know that the links within the validation dataset will not be found in the training dataset. Therefore, to reach a link found in the validation dataset, we need a minimum of 3 links, i.e. a path length of 3, in the training dataset as shown in Fig. 4 due to the bipartite nature of the graph.

Out of the 1614 links within the validation dataset which we used for the experiment, we found that 1477 required only a path length of 3 to reach using the training dataset. This shows that the vast majority of the links in the validation dataset have common neighbours in the training dataset. In Katz Index, a smaller $\beta$ value means that more emphasis is placed on paths of shorter length. Since most paths are of length 3, smaller $\beta$ value leads to better results. Another observation that can be made from Fig. 3 is the difference in results using $\beta$ of 0.02 and 0.01 is much larger than the difference in results using $\beta$ of 0.01 and 0.005. This seems to indicate that $\beta$ value affects the performance of Katz index in the DTI link prediction exponentially as its value increases. However, more experiments using a larger range of $\beta$ values need to be performed to fully confirm this.

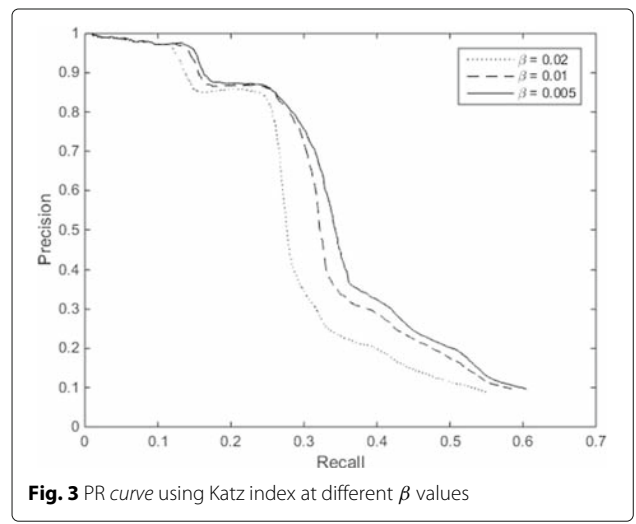

**Fig. 3** *PR curve* using Katz index at different $\beta$ values

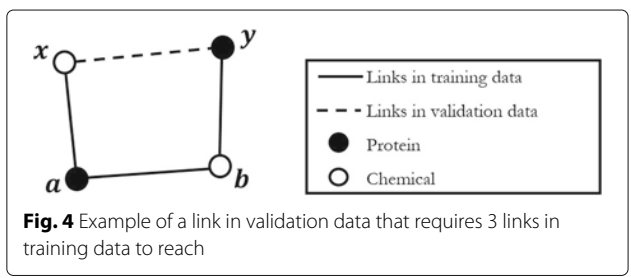

**Fig. 4** Example of a link in validation data that requires 3 links in training data to reach

**Limitations**

Although our approach has a clear advantage over the machine learning-based baseline by making better use of network topology information, it has limitations, as illustrated in Fig. 5. If a candidate link (e.g. the one between nodes x and y) happens to form a cut of the network, i.e., there are two disconnected sub-networks in the training dataset, there is no valid path suggesting the link to be a potential link, which will affect the recall. Being based entirely on network topology and not on the attributes of the individual nodes or edges similarity indices cannot predict a link between disconnected networks.

A typical example would be: when a new drug has just been invented for a new target, and no information is available on how it interacts with other targets in an exiting DTI network, the number of common neighbours between the new drug and any existing target will be zero. This means that no potential DTIs can be predicted for the new drug. In this scenario, to perform link prediction on the new drug or target, we will need to employ alternative methods that require extra knowledge about drugs and targets.

Additionally, drug-protein interactions are complex in nature. For targets with multiple distinct pockets for different drugs [29], the current method will likely fail. One possible option would be to use drug/target profiles wherever applicable to perform finer-grained DTI prediction, for instance, by splitting a single target node to a set of pocket nodes.

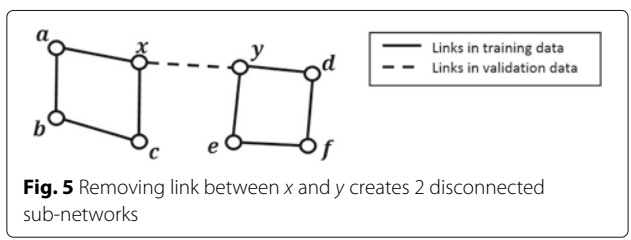

**Fig. 5** Removing link between x and y creates 2 disconnected sub-networks

**Conclusion**

*In silico* DTI prediction is aimed at assisting scientists in identification of drug-target interactions. Because the most interesting potential interactions will need to be verified in the context of costly and time-consuming laboratory experiments, reliability of DTI prediction is very important. Network-based DTI prediction is currently popular, with most methods based on machine learning (e.g. the well-known RBM). These methods tend to yield high performance when additional information about attributes of drugs, targets, and their interactions is available. However, they perform poorly when such information is limited or unavailable.

We have experimented, for the first time, whether similarity indices used e.g. in social network analysis might offer a more suitable approach to DTI prediction, when no further information is available beyond the DTI network topology. Our investigation of several similarity index -based methods shows that such an approach can indeed have a distinct advantage over the currently prevalent machine learning -based approaches when binding types and causes of interactions are unspecified. In our experiments, this advantage is particularly apparent on the top 5000 predicted links, where similarity indices have precision levels significantly higher than the RBM method. Given the time and manpower required for validation of potential interactions, it is arguably more important for the precision to be high at lower recall than at higher recall. However, when additional features about DTIs are available, it is better to use conventional machine learning-based method such as RBM. Thus, link prediction using similarity indices is not intended as a competing but rather a complementary method.

In the future our approach could be improved in various ways. One idea would be to include characteristics of the nodes and edges into similarity indices. This could be done, for example, by splitting each individual target to multiple distinct pocket nodes for finer-grained DTI prediction, which might improve the accuracy of our predicted DTIs and at the same time, overcome several current limitations of the similarity indices approach. Additionally, it would be an interesting line of research to experiment with large scale, up-to-date DTI networks with a comprehensive coverage of all approved and experimental drugs and the corresponding targets, e.g., by integrating various drug databases and literature. Any predicted DTI needs to be checked against existing literature (e.g., via PubMed search) for novelty, followed by laboratory experiments and clinical trials for further validation. We hope that this approach and its future enhancements will provide real, reliable predictions (in contrast to pseudo evaluation) that will benefit the research community and pharmaceutical industry.

## Abbreviations

CN: Common Neighbours; DTI: Drug target interactions; FN: False Negative; FP: False Positive; GUI: Graphical User Interface; MATADOR: Manually Annotated Target and Drug Online Resource; MeSH: Medical Subject Headings; PA: Preferential Attachment; PR: Precision-Recall; RBM: Restricted Boltzmann Machine; SVM: Support Vector Machine; TP: True Positive

## Acknowledgements

We would like to express our gratitude to Dr Michael Zeng and Dr Wang Yuhao, authors of the paper *Predicting drug-target interactions using restricted Boltzmann machines* for sharing the results of their experiments with us and allowing us to make a fair comparison of our method of link prediction against their method.

## Funding

We acknowledge the MRC grant MR/M013049/1.

## Availability of data and materials

The data and code for DTI prediction are available at: http://www.cl.cam.ac.uk/~yg244/16bioinfo.

## Authors' contributions

YL carried out research on the different similarity indices and also carried out the experiments by applying the similarity indices on the MATADOR database and performing 10-fold cross validation on the data. YG came up with the idea of performing pilot experiments on the MATADOR dataset to check if there are any disjoint networks within the dataset. She also provided guidance to YL by thinking of explanations to interesting observations. Anna was the supervisor of YL and was the person who came up with the idea of applying similarity indices to DTI networks for drug repositioning. All authors read and approved the final manuscript.

## Competing interests

The authors declare that they have no competing interests.

## Consent for publication

Not applicable.

## Ethics approval and consent to participate

Not applicable.